\documentclass{article}

\usepackage{arxiv}

\usepackage[utf8]{inputenc}
\usepackage[T1]{fontenc}

\usepackage{url}
\usepackage{booktabs}
\usepackage{amsfonts}
\usepackage{nicefrac}
\usepackage{microtype}
\usepackage{graphicx}
\usepackage[numbers,sort&compress]{natbib}
\usepackage{doi}

\usepackage{amsmath,amssymb}
\usepackage{bm}
\usepackage{xcolor}
\usepackage{colortbl}
\usepackage{multirow}
\usepackage{makecell}
\usepackage{tabularx}
\usepackage{enumitem}
\usepackage{float}
\usepackage{subcaption}
\usepackage{placeins}
\usepackage{pgfplots}
\usepackage{dblfloatfix}
\usepackage[ruled,linesnumbered]{algorithm2e}

\usepackage{hyperref}
\usepackage{cleveref}

\pgfplotsset{compat=1.18}
\usepgfplotslibrary{statistics}
\newcommand{\eg}{e.g.}
\newcommand{\ie}{i.e.}

\newcommand{\etc}{etc.}

\newcommand{\ours}{VGGT-CD}

\DeclareMathOperator*{\argmax}{arg\,max}
\newcommand{\vt}{\mathbf{t}}
\newcommand{\vp}{\mathbf{p}}
\newcommand{\vR}{\mathbf{R}}

\definecolor{cvprblue}{RGB}{25,100,180}
\definecolor{best}{RGB}{220,50,50}

\title{VGGT-CD: Training-Free Robust Registration for 3D Change Detection}

\date{}

\author{
Wei Zhang\\
Northwestern Polytechnical University\\
Xi’an, 710072, China\\
\texttt{zhangwei707@mail.nwpu.edu.cn}
\And
Songhua Li\\
Northwestern Polytechnical University\\
Xi’an, 710072, China\\
\texttt{lisonghua@mail.nwpu.edu.cn}
\And
Yihang Wu\\
Northwestern Polytechnical University\\
Xi’an, 710072, China\\
\texttt{wuyihang@mail.nwpu.edu.cn}
\And
Qiang Li\\
Northwestern Polytechnical University\\
Xi’an, 710072, China\\
\texttt{qiangli@nwpu.edu.cn}
\And
Qi Wang\thanks{Corresponding author.}\\
Northwestern Polytechnical University\\
Xi’an, 710072, China\\
\texttt{crabwq@nwpu.edu.cn}
}

\renewcommand{\shorttitle}{VGGT-CD}

\hypersetup{
  pdftitle={VGGT-CD: Training-Free Robust Registration for 3D Change Detection},
  pdfsubject={Computer Vision; 3D Change Detection; Point Cloud Registration},
  pdfauthor={Wei Zhang, Songhua Li, Yihang Wu, Qiang Li, Qi Wang},
  pdfkeywords={3D change detection, point cloud registration, multi-view reconstruction, VGGT, training-free method},
}

\begin{document}
\maketitle

\begin{abstract}
3D change detection from multi-view images is essential for urban monitoring, disaster assessment, and autonomous driving. However, existing methods predominantly operate in the 2D domain, where viewpoint variations are mistaken for physical changes and depth is unavailable. While visual geometry foundation models like VGGT rapidly produce dense point clouds from unposed images, independent per-epoch reconstruction encounters fundamental obstacles: unpredictable inter-epoch scale ambiguity, registration-change paradox where scene changes corrupt alignment, and pervasive edge-flying noise. To address these challenges, we present \ours{}, a training-free pipeline decoupling cross-temporal registration from dynamic-change interference. In Coarse Stage, sparse keyframe joint inference establishes a unified metric space and yields an initial $\mathrm{Sim}(3)$ prior. In Fine Stage, dense reconstructions are purified by isolating static-background correspondences. A closed-form centroid alignment refines the translation while locking scale and rotation, using a residual self-check to mathematically guarantee non-degradation. Evaluated on an 11-scene benchmark from the World Across Time dataset, \ours{} reduces Absolute Trajectory Error by 44\% outdoors and 59\% indoors. It completes registration over 6 times faster, producing high-purity 3D change maps without task-specific training.
\end{abstract}


\begin{center}
\textbf{Code:} \href{https://github.com/WZ-CS/VGGT-CD}{\texttt{https://github.com/WZ-CS/VGGT-CD}}
\end{center}

\begin{figure}[t]
  \centering
  \includegraphics[width=\linewidth]{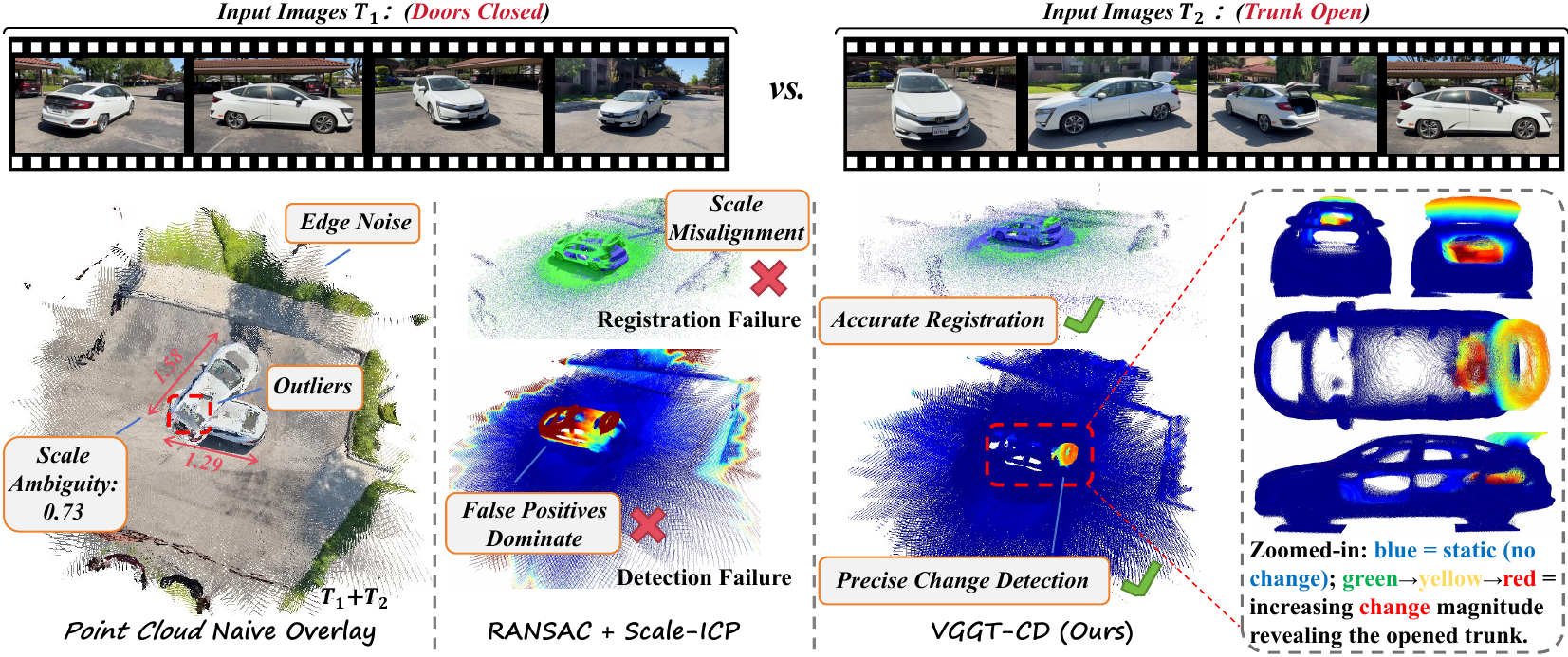}
  \caption{%
    Given bi-temporal multi-view images (\emph{doors closed} vs.\ \emph{trunk open}), independent reconstruction produces two point clouds in separate coordinate frames. Naive overlay without alignment exposes severe scale ambiguity and edge-flying noise, rendering the two epochs entirely incomparable (left). RANSAC + Scale-ICP fails to resolve the scale discrepancy, resulting in registration failure and false-positive-dominated change detection (middle left). Our VGGT-CD achieves accurate registration and precise change detection (middle right). Zoomed-in views (right): blue = static (no change); green$\to$yellow$\to$red = increasing change magnitude revealing the opened trunk.
  }
  \label{fig:teaser}
\end{figure}

\section{Introduction}\label{sec:intro}

Cross-temporal change detection plays a vital role in urban monitoring~\cite{stilla2023change,de2021change,zovathi2022point}, disaster assessment~\cite{zheng2021building,al2024integrating,canozu2025rapid}, and autonomous driving~\cite{lin2026sceneedited,albagami2025urban}. Conventional methods operate mainly in the 2D image domain~\cite{zhang2020deeply,chen2020spatial,chen2021remote}, yet remain inherently vulnerable to viewpoint variations that cause parallax-induced false positives and, more fundamentally, lack access to the depth information required to capture volumetric and structural changes. In contrast, 3D change detection compares point-cloud geometry directly, offering stronger spatial robustness and richer geometric information that is inaccessible from 2D observations alone. The recent emergence of visual geometry foundation models, including VGGT~\cite{wang2025vggt}, DUSt3R~\cite{wang2024dust3r}, and MASt3R~\cite{leroy2024grounding}, has significantly advanced this direction by enabling rapid and accurate 3D reconstruction from casual image collections: these models produce dense point clouds from unposed images in a single forward pass, greatly facilitating the acquisition of cross-temporal 3D geometry~\cite{wang2025pi,zhuo2025streaming,yuan2026infinitevggt}.

However, directly applying these foundation models
to cross-temporal change detection
faces fundamental difficulties.
An intuitive baseline is to
independently reconstruct each epoch $T_1$ and $T_2$
and then align the two point clouds
with a registration algorithm such as ICP~\cite{besl1992method}.
In practice, this approach fails
due to three intertwined problems:
(i)~\emph{scale ambiguity}:
independent reconstructions reside in separate coordinate frames
with unknown relative metric scale,
making rigid-body ($\mathrm{SE}(3)$) registration insufficient;
(ii)~the \emph{registration--change paradox}:
genuine physical changes act as massive geometric outliers
that corrupt the alignment,
while misalignment in turn produces false changes;
and (iii)~\emph{edge-flying noise}:
neural depth predictions exhibit boundary elongation artifacts
that generate spurious changes even in static regions.

To address these challenges,
we propose \ours{},
a training-free, coarse-to-fine pipeline
that decouples cross-temporal registration
from dynamic-change interference.
In \textbf{Stage~1},
we select a small number of keyframes from both epochs
and feed them into a single VGGT joint-inference pass,
implicitly establishing a unified metric space
that resolves the inter-epoch scale ambiguity
and yields an initial $\mathrm{Sim}(3)$ transform.
In \textbf{Stage~2},
we perform full dense reconstruction independently
to recover high-resolution geometric detail.
A confidence-guided purification mechanism
then filters out dynamic change regions and edge noise,
retaining only high-confidence static-background correspondences.
Based on these purified points,
a single-step SVD~\cite{umeyama2002least}
yields the refined alignment in closed form,
eliminating iterative-optimization drift entirely.
A residual self-check further guarantees
that the Fine Stage never degrades
the Coarse result.

Experiments on public multi-temporal datasets with ground-truth camera trajectories show that \ours{} achieves the best registration accuracy across all 11 evaluated scenes, reducing ATE by up to 59\% over the strongest baseline while completing registration over $6\times$ faster.

Our contributions are as follows:
\begin{itemize}

\item
We propose \ours{}, a training-free, coarse-to-fine 3D change detection pipeline. By exploiting the joint-inference capability of visual geometry foundation models to establish a unified metric prior, our method is, to the best of our knowledge, the first to perform end-to-end 3D change detection from unposed multi-view images without traditional SfM or explicit feature matching.

\item We design a reliability-guided purification mechanism with a decoupled translation refinement and a residual self-check, providing a \emph{monotonicity guarantee} that the Fine Stage provably never degrades the Coarse alignment.

\item We establish a rigorous cross-temporal 3D registration evaluation protocol on the public World Across Time dataset, covering 11 scenes with ground-truth trajectories. Extensive experiments demonstrate state-of-the-art registration accuracy and over $6\times$ speedups, yielding high-purity 3D change maps.

\end{itemize}

\section{Related Work}\label{sec:related}

\paragraph{\textbf{3D point cloud change detection.}}
Traditional change detection methods
predominantly operate in the 2D image domain,
identifying changes through feature differencing~\cite{tewkesbury2015critical,singh1989review}.
However, 2D approaches are sensitive
to illumination variations and viewpoint parallax,
and cannot capture volumetric or structural changes.
To overcome these limitations,
3D change detection has attracted growing attention.
Existing 3D methods typically rely on
LiDAR scans~\cite{wehr1999airborne}
or time-consuming SfM pipelines~\cite{snavely2006photo}
to obtain the input point clouds,
and then employ distance metrics
or deep networks to extract change features.
PGN3DCD~\cite{zhan2024pgn3dcd}
introduces non-parametric priors
to guide multi-temporal point cloud differencing,
while Living Scenes~\cite{zhu2024living}
attempts multi-object reconstruction
and relocalization in dynamic environments.
These methods, however, depend heavily on
pre-aligned point clouds
and cannot avoid the slow multi-view geometric optimization.
In contrast, \ours{} completely bypasses traditional SfM
and, for the first time,
leverages an unposed visual geometry foundation model
for end-to-end cross-temporal 3D change detection.

\paragraph{\textbf{Visual geometry foundation models.}}
The 3D vision community has recently witnessed
a surge of foundation models
for geometry estimation~\cite{wang2024dust3r,leroy2024grounding,wang2025pi}.
DUSt3R~\cite{wang2024dust3r}
and MASt3R~\cite{leroy2024grounding}
reformulate multi-view stereo~\cite{yao2018mvsnet,gu2020cascade,zhang2024visual,11068977,11112633,mao2024sdl,zhang2024edge,li2023hierarchical,peng2018red,peng2022rethinking}
as a pointmap regression task,
producing dense 3D reconstructions
from sparse image pairs without camera pose priors.
VGGT~\cite{wang2025vggt}
further unifies this framework
by jointly predicting
camera extrinsics, intrinsics, depth maps,
and 3D feature tracks
in a single feed-forward pass.
While these models achieve impressive per-scene results,
applying them independently to each epoch
of a cross-temporal task
inevitably introduces random drift
in metric scale and local coordinate frames.
\ours{} addresses this by introducing
a sparse joint-inference mechanism
that resolves the inherent scale ambiguity
of independent feed-forward reconstruction.

\paragraph{\textbf{Robust point cloud registration.}}
Point cloud registration is a prerequisite
for comparing cross-temporal 3D geometry.
The classical ICP algorithm~\cite{besl1992method}
is prone to local minima
when facing low overlap or abundant outliers.
Fast Global Registration (FGR)~\cite{zhou2016fast}
and RANSAC-based pipelines
improve robustness through global search,
while learning-based methods such as
GeoTransformer~\cite{qin2023geotransformer}
predict correspondences via geometric transformers,
and Dynamic Cues-Assisted Transformer~\cite{chen2024dynamic}
leverages dynamic prompts
to disambiguate feature matching.
SGHR~\cite{wang2023robust}
filters erroneous matches in the pose graph
through history reweighting.
Despite these advances,
in cross-temporal change detection
the genuine physical changes
constitute structured, large-scale geometric outliers
that overwhelm even the most robust estimators,
giving rise to the ``registration--change paradox.''
\ours{} breaks this paradox
by pre-isolating dynamic changes and edge noise
through a reliability-guided purification strategy
before applying the classical Umeyama
algorithm~\cite{umeyama2002least}
in closed form,
circumventing iterative registration drift entirely
and achieving over $6\times$ speedups.

\section{Method}\label{sec:method}

Given two sets of multi-view images
$\mathcal{I}^{T_1}{=}\{I^{T_1}_1,\dots,I^{T_1}_{N_1}\}$ and
$\mathcal{I}^{T_2}{=}\{I^{T_2}_1,\dots,I^{T_2}_{N_2}\}$
captured at two different time epochs $T_1$ and $T_2$,
our goal is to align the corresponding 3D reconstructions
into a shared metric coordinate system and detect geometric changes between them.
The core challenge arises from the fact that
feed-forward reconstruction models~\cite{wang2025vggt}
recover each epoch independently,
producing point clouds
$\mathcal{P}^{T_1}$ and $\mathcal{P}^{T_2}$
that reside in \emph{separate} coordinate frames
with unknown relative scale $s$, rotation $\vR$, and translation $\vt$,
\ie a seven-degree-of-freedom $\mathrm{Sim}(3)$ ambiguity.

\begin{figure*}[t]
\centering
\includegraphics[width=\linewidth]{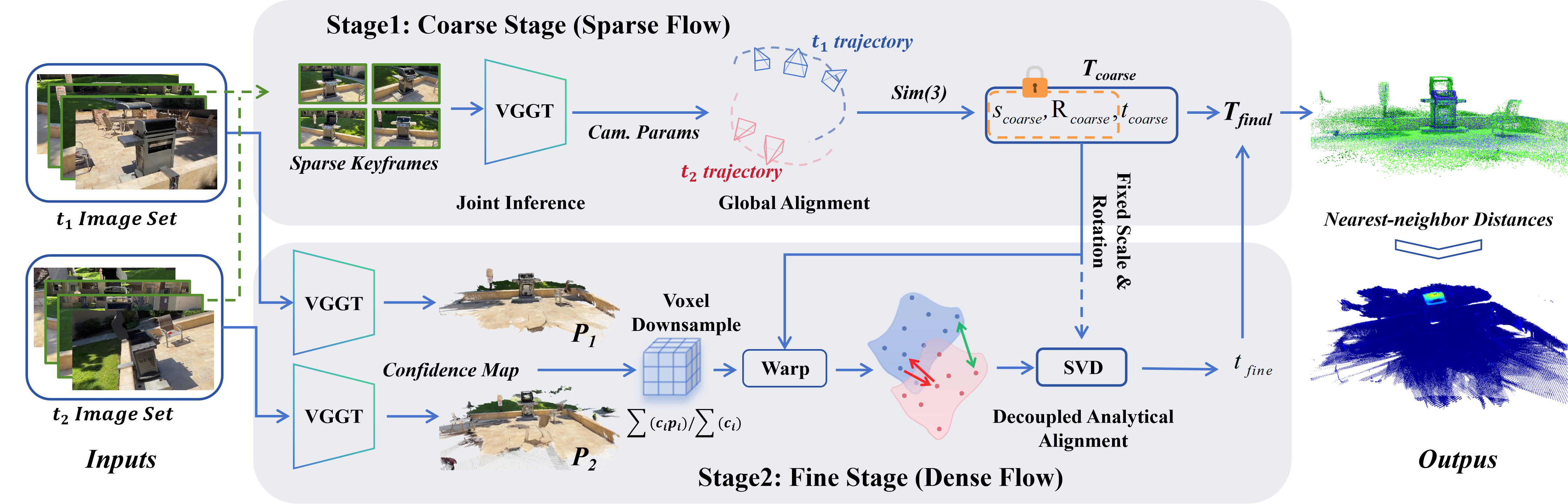}
\caption{\textbf{Overall architecture of the proposed VGGT-CD pipeline.} Given unposed image sets from two temporal states ($T_1$ and $T_2$), our training-free system operates in a decoupled, coarse-to-fine manner. \textbf{(Top) Coarse Stage:} A sparse subset of keyframes undergoes joint inference to establish a unified metric space. By aligning the implicitly reconstructed camera frustums, we extract a reliable global prior, rigidly locking the scale ($s_{coarse}$) and rotation ($\mathbf{R}_{coarse}$) from the Coarse Stage. \textbf{(Bottom) Fine Stage:} Independent dense point clouds are first processed via voxel-based spatial equalization to reduce foreground density bias. After projecting $T_1$ with the coarse prior, a nearest-neighbor matching mechanism isolates true physical changes from the static background. Finally, a decoupled analytical alignment solves purely for the translation offset ($\mathbf{t}_{fine}$), preventing the lever-arm effect from local depth noise. \textbf{(Right):} The aligned point clouds yield the final 3D volumetric change field.}
\label{fig:pipeline}
\end{figure*}

We propose \ours{}, a coarse-to-fine pipeline (Fig.~\ref{fig:pipeline})
that resolves this ambiguity without traditional Structure-from-Motion
or explicit feature matching across epochs.
Stage~1 (\emph{Coarse}, Sec.~\ref{sec:coarse})
establishes a metric-scale prior through sparse joint inference;
Stage~2 (\emph{Fine}, Sec.~\ref{sec:fine})
refines the translation via dense point-cloud purification
with a decoupled, degradation-free design.
Sec.~\ref{sec:cd} describes how the aligned point clouds
are compared to produce the final change map.

\subsection{Per-Epoch Dense Reconstruction}\label{sec:recon}

For each epoch $T_k$ ($k \!\in\! \{1,2\}$),
we feed all $N_k$ images into VGGT~\cite{wang2025vggt}
in a single forward pass.
VGGT jointly predicts, for every frame~$i$,
camera extrinsics $[\vR_i \mid \vt_i] \!\in\! \mathrm{SE}(3)$,
intrinsics $\mathbf{K}_i$,
per-pixel depth $D_i$,
and a depth confidence map $c_i \!\in\! [0,1]$.
Dense 3D points are obtained by back-projecting each pixel
using the predicted depth and camera parameters:
\begin{equation}\label{eq:backproj}
  \vp_{i,u} \;=\; \vR_i^{-1}
  \!\left( D_i(u)\,\mathbf{K}_i^{-1}\tilde{u} - \vt_i \right),
\end{equation}
where $\tilde{u}$ denotes the homogeneous pixel coordinate.
Aggregating over all frames yields the per-epoch
point cloud $\mathcal{P}^{T_k}$ with associated confidence scores.

Because each epoch is reconstructed \emph{independently},
the resulting coordinate frames are arbitrary:
$\mathcal{P}^{T_1}$ and $\mathcal{P}^{T_2}$ are neither scale-consistent
nor pose-aligned, precluding direct comparison.

\subsection{Coarse Registration via Sparse Joint Inference}
\label{sec:coarse}

The key insight behind our coarse stage is that
VGGT's joint inference on a mixed set of images from \emph{both} epochs
implicitly places all frames into a single metric coordinate system,
thereby resolving the inter-epoch $\mathrm{Sim}(3)$ ambiguity.

\paragraph{\textbf{Keyframe selection.}}
Rather than feeding all $N_1 {+} N_2$ frames jointly
(which would exceed GPU memory for typical sequences),
we select a small representative subset.
Specifically, from each epoch we sample $K$ keyframes
using \emph{Farthest Point Sampling}~(FPS) in temporal index space.
Let $\{1, \dots, N_k\}$ be the frame indices for epoch $T_k$.
We first select the start frame $k_1 {=} 1$,
then iteratively choose:
\begin{equation}\label{eq:fps}
  k_{j+1} = \argmax_{i \,\in\, \{1,\dots,N_k\} \setminus \mathcal{K}}
  \;\min_{k \in \mathcal{K}} |i - k|,
\end{equation}
where $\mathcal{K}$ is the already-selected set.
This yields a temporally maximally-spread subset
$\mathcal{K}^{T_k} {=} \{k_1, \dots, k_K\}$,
ensuring broad viewpoint coverage with minimal redundancy.
Note that when the input frames are ordered sequentially,
Eq.~\eqref{eq:fps} approximately uniform temporal sampling
with indices $\{1, \lceil\frac{N_k}{K}\rceil, \lceil\frac{2 N_k}{K}\rceil, \dots\}$,
but FPS naturally handles non-uniformly ordered inputs.

\paragraph{\textbf{Joint inference.}}
We form a mixed batch
$\mathcal{B} = \{I^{T_1}_{k}\}_{k \in \mathcal{K}^{T_1}}
  \cup \{I^{T_2}_{k}\}_{k \in \mathcal{K}^{T_2}}$
of $2K$ images and feed it to VGGT in a single forward pass.
The resulting predictions
$\hat{\mathcal{P}}^{\mathrm{joint}}$
are expressed in a unified metric frame that is shared
across both epochs.

\paragraph{\textbf{Sim(3) estimation.}}
For each epoch $T_k$, we now have two representations
of the same keyframe point cloud:
(i)~the per-epoch reconstruction $\mathcal{P}^{T_k}_{\mathrm{kf}}$
(from Sec.~\ref{sec:recon}),
and
(ii)~the joint-inference reconstruction
$\hat{\mathcal{P}}^{\mathrm{joint}}_{k}$.
These form dense correspondences from which we estimate
the $\mathrm{Sim}(3)$ transform
$(s_k, \vR_k, \vt_k)$
that maps the per-epoch frame into the joint frame,
using the closed-form Umeyama algorithm~\cite{umeyama2002least}.
To enhance robustness, we retain only points whose
depth confidence exceeds the median value
and cap the correspondence count at $M{=}5000$ via random subsampling.

Given the two per-epoch transforms
$(s_1, \vR_1, \vt_1)$ and $(s_2, \vR_2, \vt_2)$,
the relative $\mathrm{Sim}(3)$ that directly maps
$\mathcal{P}^{T_1}$ into the coordinate frame of $\mathcal{P}^{T_2}$
is obtained by composing the two:
\begin{equation}\label{eq:sim3_compose}
  s = \frac{s_1}{s_2}, \qquad
  \vR = \vR_2^{\!\top} \vR_1, \qquad
  \vt = \frac{1}{s_2}\,\vR_2^{\!\top}\!\left(\vt_1 - \vt_2\right).
\end{equation}
The aligned point cloud is then
$\tilde{\mathcal{P}}^{T_1} = \{s\,\vR\,\vp + \vt \mid \vp \!\in\! \mathcal{P}^{T_1}\}$.

\subsection{Fine Registration via Decoupled Translation Refinement}
\label{sec:fine}

The coarse stage uses only $K$ keyframes per epoch
(typically $K{=}5$),
limiting the geometric information available for alignment.
When $K$ is small or the selected frames observe limited scene overlap,
the resulting translation $\vt$ may be sub-optimal.
Stage~2 leverages the \emph{full} dense point clouds from both epochs
to refine the alignment, while carefully avoiding the introduction of new errors.

\begin{algorithm}[t]
  \caption{Stage 2: Decoupled Translation Refinement}
  \label{alg:fine}
  \SetKwInOut{Input}{Input}
  \SetKwInOut{Output}{Output}

  \Input{Dense point clouds $\bar{\mathcal{P}}^{T_1}$, $\bar{\mathcal{P}}^{T_2}$ (after voxel downsampling); coarse transform $(s, \vR, \vt)$}
  \Output{Refined translation $\vt_{\mathrm{final}}$}

  \BlankLine
  \tcp{1. Coarse Alignment \& Distance Computation}
  Apply coarse transform: $\tilde{\vp}_i \leftarrow s\,\vR\,\vp_i + \vt$, $\forall\, \vp_i \in \bar{\mathcal{P}}^{T_1}$\;
  Compute nearest-neighbor distances: $d_i \leftarrow \min_{\vp'} \| \tilde{\vp}_i - \vp' \|$\;

  \BlankLine
  \tcp{2. Static Set Extraction}
  Extract static set: $\mathcal{S} \leftarrow \{i \mid d_i < 3 \cdot \mathrm{median}(\{d_j\})\}$\;
  \If{$|\mathcal{S}| < 100$}{
    \tcp{Too few static points; keep coarse}
    \Return $\vt$\;
  }

  \BlankLine
  \tcp{3. Refined Translation Computation}
  Compute refined translation: $\vt^* \leftarrow \frac{1}{|\mathcal{S}|}\sum_{i \in \mathcal{S}} (\vp'_{\mathrm{nn}(i)} - s\,\vR\,\vp_i)$\;

  \BlankLine
  \tcp{4. Residual Self-Check}
  \eIf{$\mathrm{median}(\{d^*_j\}) < \mathrm{median}(\{d_j\})$}{
    \tcp{Fine improves alignment}
    \Return $\vt^*$\;
  }{
    \tcp{Revert to coarse}
    \Return $\vt$\;
  }
\end{algorithm}

\paragraph{\textbf{Confidence-weighted voxel downsampling.}}
We first filter both point clouds by retaining only points
with depth confidence above the median,
then apply voxel grid downsampling with an adaptive voxel size
$\delta = L / G$,
where $L$ is the 99th-percentile spatial extent
and $G{=}200$ is a fixed grid resolution.
Within each voxel, the point with the highest confidence score
is retained as the representative.
This produces compact, high-quality subsets
$\bar{\mathcal{P}}^{T_1}$ and $\bar{\mathcal{P}}^{T_2}$
of approximately $10^4$--$10^5$ points each.

\paragraph{\textbf{Dynamic change-region purification.}}
Using the coarse transform from Eq.~\eqref{eq:sim3_compose},
we align $\bar{\mathcal{P}}^{T_1}$ to the coordinate frame
of $\bar{\mathcal{P}}^{T_2}$
and compute, for each point in $\tilde{\bar{\mathcal{P}}}^{T_1}$,
the nearest-neighbor distance to $\bar{\mathcal{P}}^{T_2}$:
\begin{equation}\label{eq:nn_dist}
  d_i = \min_{\vp' \in \bar{\mathcal{P}}^{T_2}}
  \left\| (s\,\vR\,\vp_i + \vt) - \vp' \right\|_2,
  \quad \vp_i \in \bar{\mathcal{P}}^{T_1}.
\end{equation}
Points with small $d_i$ correspond to the \emph{static} background,
while large $d_i$ indicates genuine 3D changes
(\eg objects that appeared, disappeared, or moved).
We define the static set as:
\begin{equation}\label{eq:static}
  \mathcal{S} = \left\{
    i \;\middle|\;
    d_i < \alpha \cdot \mathrm{median}(\{d_j\}_j)
  \right\},
\end{equation}
where $\alpha{=}3.0$ provides an adaptive, scene-dependent threshold.

\paragraph{\textbf{Decoupled translation refinement.}}
A na\"ive approach would re-estimate
the full $\mathrm{Sim}(3)$ from the static correspondences.
However, small errors in the re-estimated rotation
can produce large translation errors at distant points
through a \emph{lever-arm} effect---
a rotation error of $\Delta\theta$ causes
a positional displacement of $r \cdot \Delta\theta$
at radius~$r$ from the centroid.
We therefore adopt a \emph{decoupled} strategy:
$s$ and $\vR$ are \textbf{locked} at their coarse-stage values,
and only the translation $\vt$ is updated.
The optimal refined translation is given by
the centroid difference of the purified static correspondences:
\begin{equation}\label{eq:refine_t}
  \vt^{*} = \frac{1}{|\mathcal{S}|}
  \sum_{i \in \mathcal{S}}
  \left(
    \vp'_{\mathrm{nn}(i)} - s\,\vR\,\vp_i
  \right),
\end{equation}
where $\vp'_{\mathrm{nn}(i)} \in \bar{\mathcal{P}}^{T_2}$
is the nearest neighbor of the $i$-th aligned source point.

This formulation has a favorable theoretical property:
since we optimize a single three-dimensional vector
from $|\mathcal{S}|$ ($\gg 3$) correspondences,
the problem is massively over-determined,
yielding a statistically robust estimate.

\paragraph{\textbf{Residual self-check.}}
To further guarantee that the fine stage never degrades
the coarse result, we introduce a \emph{residual self-check}.
After computing $\vt^*$,
we evaluate the median nearest-neighbor residual
over \emph{all} downsampled points (not just the static subset)
for both the coarse translation $\vt$ and the refined $\vt^*$:
\begin{equation}\label{eq:selfcheck}
  \vt_{\mathrm{final}} =
  \begin{cases}
    \vt^{*}, & \text{if } \mathrm{med}(\{d^*_j\}_j) < \mathrm{med}(\{d_j\}_j), \\[3pt]
    \vt,     & \text{otherwise},
  \end{cases}
\end{equation}
where $d^*_j$ and $d_j$ denote the nearest-neighbor distances
under $\vt^*$ and $\vt$, respectively.
This mechanism ensures that our pipeline satisfies
the \emph{monotonicity guarantee}:
$\mathrm{ATE}_{\mathrm{Full}} \leq \mathrm{ATE}_{\mathrm{Coarse}}$
for every scene and keyframe configuration.
The overall fine stage is summarized in Algorithm~\ref{alg:fine}.

\subsection{3D Change Detection}\label{sec:cd}

With the aligned point clouds
$\tilde{\mathcal{P}}^{T_1}$ (in the frame of $\mathcal{P}^{T_2}$),
we detect 3D changes by computing bidirectional
nearest-neighbor distances.
For each point $\vp \in \tilde{\mathcal{P}}^{T_1}$,
the change score is defined as:
\begin{equation}\label{eq:cd}
  \Delta(\vp) = \min_{\vp' \in \mathcal{P}^{T_2}}
  \|\vp - \vp'\|_2.
\end{equation}
Symmetrically, we compute changes from $T_2$ to $T_1$.
A point is classified as \emph{changed} if
$\Delta(\vp) > \tau_{\mathrm{cd}}$,
where $\tau_{\mathrm{cd}}$ is set proportionally
to the scene extent.
The union of forward and backward change detections
produces the final change map,
which can be projected back to individual images
for per-view change visualization.

\section{Experiments}\label{sec:exp}

\subsection{Experimental Setup}\label{sec:setup}

\paragraph{\textbf{Dataset and benchmark.}}
Existing 3D change detection datasets predominantly provide
only 2D masks for evaluation,
lacking rigorous metrics for cross-temporal geometric alignment.
To address this,
we establish a 3D registration evaluation protocol on the WAT dataset~\cite{cai2023clnerf},
which contains real-world scenes captured at different time epochs
with significant physical changes
(moved furniture, opened doors, added/removed objects, \etc).
Our benchmark comprises \textbf{11 scenes}:
5~outdoor (\textit{Car}, \textit{Community}, \textit{Grill},
\textit{Spa}, \textit{Street})
and 6~indoor (\textit{Breville}, \textit{Dyson}, \textit{Kitchen},
\textit{Living Room}, \textit{Mac}, \textit{Ninja}).
These scenes exhibit highly diverse challenges,
including extreme scale variations in outdoor environments,
textureless walls and reflective surfaces indoors,
and large-scale physical changes across epochs.
We extract the ground-truth camera trajectories
from the static-background COLMAP reconstruction
provided by the WAT dataset,
enabling absolute evaluation of cross-temporal alignment quality.

\paragraph{\textbf{Baselines.}}
We compare against representative combinations of
state-of-the-art feed-forward reconstruction
and robust point cloud registration.
All baselines use VGGT~\cite{wang2025vggt}
as the shared reconstruction backbone,
followed by different registration strategies:
\textbf{(i)}~\emph{VGGT Only}:
independent per-epoch reconstruction without any cross-temporal alignment
(the identity transform serves as the baseline);
\textbf{(ii)}~\emph{VGGT + FGR}~\cite{zhou2016fast}:
Fast Global Registration with scale estimation
on the dense point clouds;
\textbf{(iii)}~\emph{VGGT + RANSAC+ICP}:
RANSAC-based global registration followed by
scale-aware ICP refinement;
\textbf{(iv)}~\emph{VGGT + Scale+ICP}:
relative scale estimation via correspondence distance ratios
followed by standard ICP alignment;
\textbf{(v)}~\emph{GeoTransformer}~\cite{qin2023geotransformer}:
a learning-based robust point cloud registration method
that predicts correspondences via geometric transformers.
This set of baselines comprehensively covers the mainstream paradigm
of ``independent reconstruction + post-hoc registration,''
allowing us to isolate the contribution
of our joint-inference approach.

\paragraph{\textbf{Evaluation metrics.}}
We evaluate cross-temporal registration quality
using two complementary and widely adopted metrics:
\emph{Absolute Trajectory Error}~(ATE, in meters),
which measures global alignment accuracy
of the combined camera trajectory from both epochs,
and \emph{Relative Translation Error}~(RTE, in meters),
which captures local frame-to-frame translation consistency.
We additionally report the registration time (in seconds)
for each method,
excluding the shared VGGT per-epoch inference time
which is identical across all methods.

\begin{table*}[t]
  \centering
    \caption{%
    \textbf{Cross-temporal registration results on outdoor scenes.}
    ATE and RTE in meters (lower is better).
    T denotes registration time in seconds
    (excluding the shared VGGT per-epoch inference).
    \colorbox{cvprblue!35}{\scriptsize\color{white}1st}\,/\,%
    \colorbox{cvprblue!18}{\scriptsize 2nd}\,/\,%
    \colorbox{cvprblue!8}{\scriptsize 3rd}
    indicate the top three results per scene.
  }
  \label{tab:outdoor}
  \setlength{\tabcolsep}{5pt}
  \renewcommand{\arraystretch}{1.15}
  \small
  \resizebox{\textwidth}{!}{%
  \begin{tabular}{@{}l
    ccc
    ccc
    ccc
    ccc
    ccc
    @{}}
    \toprule
    & \multicolumn{3}{c}{\textit{Car}}
    & \multicolumn{3}{c}{\textit{Community}}
    & \multicolumn{3}{c}{\textit{Grill}}
    & \multicolumn{3}{c}{\textit{Spa}}
    & \multicolumn{3}{c}{\textit{Street}} \\
    \cmidrule(lr){2-4} \cmidrule(lr){5-7} \cmidrule(lr){8-10}
    \cmidrule(lr){11-13} \cmidrule(lr){14-16}
    Method
    & ATE$\downarrow$ & RTE$\downarrow$ & T$\downarrow$
    & ATE$\downarrow$ & RTE$\downarrow$ & T$\downarrow$
    & ATE$\downarrow$ & RTE$\downarrow$ & T$\downarrow$
    & ATE$\downarrow$ & RTE$\downarrow$ & T$\downarrow$
    & ATE$\downarrow$ & RTE$\downarrow$ & T$\downarrow$ \\
    \midrule
    VGGT Only
    & 3.373 & 0.687 & --
    & 1.703 & 0.449 & --
    & 0.716 & 0.151 & --
    & 1.661 & \cellcolor{cvprblue!8} 0.281 & --
    & 1.564 & \cellcolor{cvprblue!8} 0.263 & -- \\
    FGR
    & 0.405 & \cellcolor{cvprblue!8} 0.140 & 586
    & 0.323 & 0.182 & 837
    & 0.647 & 0.151 & 166
    & 0.616 & 0.306 & 48
    & 1.083 & 0.388 & 94 \\
    RANSAC+ICP
    & \cellcolor{cvprblue!8} 0.397 & \cellcolor{cvprblue!8} 0.141 & \cellcolor{cvprblue!8} 38
    & 0.330 & \cellcolor{cvprblue!8} 0.149 & 95
    & \cellcolor{cvprblue!18} 0.445 & 0.151 & \cellcolor{cvprblue!18} 41
    & \cellcolor{cvprblue!18} 0.566 & \cellcolor{cvprblue!18} 0.301 & \cellcolor{cvprblue!8} 17
    & 1.044 & 0.402 & \cellcolor{cvprblue!18} 34 \\
    Scale+ICP
    & 0.631 & 0.143 & \cellcolor{cvprblue!18} 50
    & \cellcolor{cvprblue!18} 0.079 & \cellcolor{cvprblue!18} 0.068 & 95
    & 0.509 & 0.151 & \cellcolor{cvprblue!8} 39
    & \cellcolor{cvprblue!18} 0.566 & \cellcolor{cvprblue!8} 0.302 & \cellcolor{cvprblue!18} 10
    & \cellcolor{cvprblue!18} 0.972 & 0.424 & \cellcolor{cvprblue!8} 12 \\
    GeoTransformer
    & \cellcolor{cvprblue!18} 0.381 & \cellcolor{cvprblue!18} 0.140 & 76
    & \cellcolor{cvprblue!8} 0.150 & \cellcolor{cvprblue!35} 0.055 & 115
    & \cellcolor{cvprblue!8} 0.461 & 0.151 & 77
    & 0.610 & 0.305 & 26
    & \cellcolor{cvprblue!8} 1.030 & 0.444 & 58 \\
    \midrule
    VGGT-CD (Ours)
    & \cellcolor{cvprblue!35} 0.040 & 0.141 & \cellcolor{cvprblue!35} 3.2
    & \cellcolor{cvprblue!35} 0.018 & \cellcolor{cvprblue!35} 0.022 & \cellcolor{cvprblue!35} 5.3
    & \cellcolor{cvprblue!35} 0.037 & 0.156 & \cellcolor{cvprblue!35} 4.4
    & \cellcolor{cvprblue!35} 0.553 & \cellcolor{cvprblue!35} 0.294 & \cellcolor{cvprblue!35} 5.3
    & \cellcolor{cvprblue!35} 0.834 & \cellcolor{cvprblue!18} 0.264 & \cellcolor{cvprblue!35} 5.3 \\
    \bottomrule
  \end{tabular}%
  }
\end{table*}

\begin{table*}[t]
  \centering
  \caption{%
    \textbf{Cross-temporal registration results on indoor scenes.}
    Same format as Table~\ref{tab:outdoor}.
    Indoor scenes feature textureless walls, reflective surfaces,
    and large-scale furniture rearrangements.
  }
  \label{tab:indoor}
  \setlength{\tabcolsep}{4pt}
  \renewcommand{\arraystretch}{1.15}
  \small
  \resizebox{\textwidth}{!}{%
  \begin{tabular}{@{}l
    ccc
    ccc
    ccc
    ccc
    ccc
    ccc
    @{}}
    \toprule
    & \multicolumn{3}{c}{\textit{Breville}}
    & \multicolumn{3}{c}{\textit{Dyson}}
    & \multicolumn{3}{c}{\textit{Kitchen}}
    & \multicolumn{3}{c}{\textit{Living Room}}
    & \multicolumn{3}{c}{\textit{Mac}}
    & \multicolumn{3}{c}{\textit{Ninja}} \\
    \cmidrule(lr){2-4} \cmidrule(lr){5-7} \cmidrule(lr){8-10}
    \cmidrule(lr){11-13} \cmidrule(lr){14-16} \cmidrule(lr){17-19}
    Method
    & ATE$\downarrow$ & RTE$\downarrow$ & T$\downarrow$
    & ATE$\downarrow$ & RTE$\downarrow$ & T$\downarrow$
    & ATE$\downarrow$ & RTE$\downarrow$ & T$\downarrow$
    & ATE$\downarrow$ & RTE$\downarrow$ & T$\downarrow$
    & ATE$\downarrow$ & RTE$\downarrow$ & T$\downarrow$
    & ATE$\downarrow$ & RTE$\downarrow$ & T$\downarrow$ \\
    \midrule
    VGGT Only
    & 2.547 & 1.010 & --
    & 0.947 & 0.283 & --
    & 0.855 & 0.076 & --
    & 0.743 & \cellcolor{cvprblue!8} 0.281 & --
    & 0.349 & \cellcolor{cvprblue!8} 0.047 & --
    & 0.634 & \cellcolor{cvprblue!18} 0.139 & -- \\
    FGR
    & 1.626 & 0.417 & 705
    & 2.205 & 0.709 & 805
    & \cellcolor{cvprblue!18} 0.065 & \cellcolor{cvprblue!8} 0.049 & \cellcolor{cvprblue!18} 10
    & 0.839 & 0.282 & 199
    & \cellcolor{cvprblue!8} 0.112 & \cellcolor{cvprblue!35} 0.046 & 172
    & 2.556 & 0.665 & 4608 \\
    RANSAC+ICP
    & \cellcolor{cvprblue!8} 1.419 & \cellcolor{cvprblue!35} 0.200 & 107
    & \cellcolor{cvprblue!8} 1.009 & \cellcolor{cvprblue!35} 0.283 & 98
    & 0.075 & \cellcolor{cvprblue!8} 0.053 & \cellcolor{cvprblue!8} 11
    & \cellcolor{cvprblue!8} 0.674 & \cellcolor{cvprblue!35} 0.276 & \cellcolor{cvprblue!8} 40
    & 0.113 & \cellcolor{cvprblue!35} 0.046 & \cellcolor{cvprblue!18} 28
    & 1.269 & \cellcolor{cvprblue!35} 0.138 & 204 \\
    Scale+ICP
    & \cellcolor{cvprblue!18} 0.475 & \cellcolor{cvprblue!8} 0.208 & \cellcolor{cvprblue!18} 47
    & \cellcolor{cvprblue!18} 0.596 & \cellcolor{cvprblue!8} 0.292 & \cellcolor{cvprblue!18} 51
    & 0.085 & \cellcolor{cvprblue!18} 0.047 & \cellcolor{cvprblue!35} 6
    & \cellcolor{cvprblue!18} 0.523 & \cellcolor{cvprblue!35} 0.276 & \cellcolor{cvprblue!18} 17
    & 0.234 & 0.054 & \cellcolor{cvprblue!8} 17
    & \cellcolor{cvprblue!35} 0.076 & 0.163 & \cellcolor{cvprblue!18} 19 \\
    GeoTransformer
    & \cellcolor{cvprblue!8} 1.328 & \cellcolor{cvprblue!35} 0.200 & 1038
    & 1.515 & \cellcolor{cvprblue!18} 0.290 & 509
    & \cellcolor{cvprblue!8} 0.068 & \cellcolor{cvprblue!8} 0.051 & 16
    & 0.824 & \cellcolor{cvprblue!18} 0.281 & 155
    & \cellcolor{cvprblue!18} 0.108 & \cellcolor{cvprblue!35} 0.046 & \cellcolor{cvprblue!8} 36
    & 1.060 & \cellcolor{cvprblue!8} 0.140 & 346 \\
    \midrule
    VGGT-CD (Ours)
    & \cellcolor{cvprblue!35} 0.041 & 0.217 & \cellcolor{cvprblue!35} 3.6
    & \cellcolor{cvprblue!35} 0.110 & 0.328 & \cellcolor{cvprblue!35} 3.9
    & \cellcolor{cvprblue!35} 0.055 & \cellcolor{cvprblue!35} 0.045 & \cellcolor{cvprblue!35} 3.1
    & \cellcolor{cvprblue!35} 0.441 & 0.285 & \cellcolor{cvprblue!35} 5.6
    & \cellcolor{cvprblue!35} 0.057 & 0.049 & \cellcolor{cvprblue!35} 6.0
    & \cellcolor{cvprblue!18} 0.111 & 0.166 & \cellcolor{cvprblue!35} 3.3 \\
    \bottomrule
  \end{tabular}%
  }
\end{table*}

\subsection{Quantitative Evaluation}\label{sec:quant}

Tables~\ref{tab:outdoor} and~\ref{tab:indoor}
present per-scene results on the outdoor and indoor benchmarks,
respectively.

\paragraph{\textbf{Outdoor scenes (Table~\ref{tab:outdoor}).}}
VGGT-CD achieves the lowest ATE in all five scenes.
On \textit{Car}, \textit{Community}, and \textit{Grill},
our method attains ATE below 0.04\,m,
outperforming the per-scene strongest competitor by
$9.5{\times}$, $4.4{\times}$, and $12.0{\times}$, respectively.
Even on the more challenging \textit{Spa} and \textit{Street} scenes,
where all methods struggle due to extreme depth range and sparse overlap,
VGGT-CD still achieves the best accuracy
(0.55\,m and 0.83\,m).
On average, VGGT-CD reduces outdoor ATE
from 0.53\,m (best single-method baseline, GeoTransformer)
to 0.30\,m,
a \textbf{44\%} relative improvement.

\paragraph{\textbf{Indoor scenes (Table~\ref{tab:indoor}).}}
The advantages are even more pronounced indoors.
On \textit{Breville}, most baselines are severely affected
by the scale ambiguity (ATE $\geq$1.3\,m),
with Scale+ICP performing best among them at 0.48\,m,
while VGGT-CD achieves 0.04\,m,
an $11.6{\times}$ improvement.
On \textit{Dyson}, \textit{Kitchen}, \textit{Living Room},
and \textit{Mac}, VGGT-CD consistently ranks first,
with ATE ranging from 0.06 to 0.44\,m.
A notable exception is \textit{Ninja},
where Scale+ICP achieves 0.076\,m,
outperforming our method (0.111\,m).
We attribute this to the scene's relatively small scale variation
($s \approx 1.0$), which favors direct scale-aware ICP.
Despite this single case,
VGGT-CD achieves the best average indoor ATE (0.14\,m)
over the strongest baseline (Scale+ICP, 0.33\,m),
a \textbf{59\%} relative improvement.

\begin{figure*}[t]
  \centering
  \includegraphics[width=\linewidth]{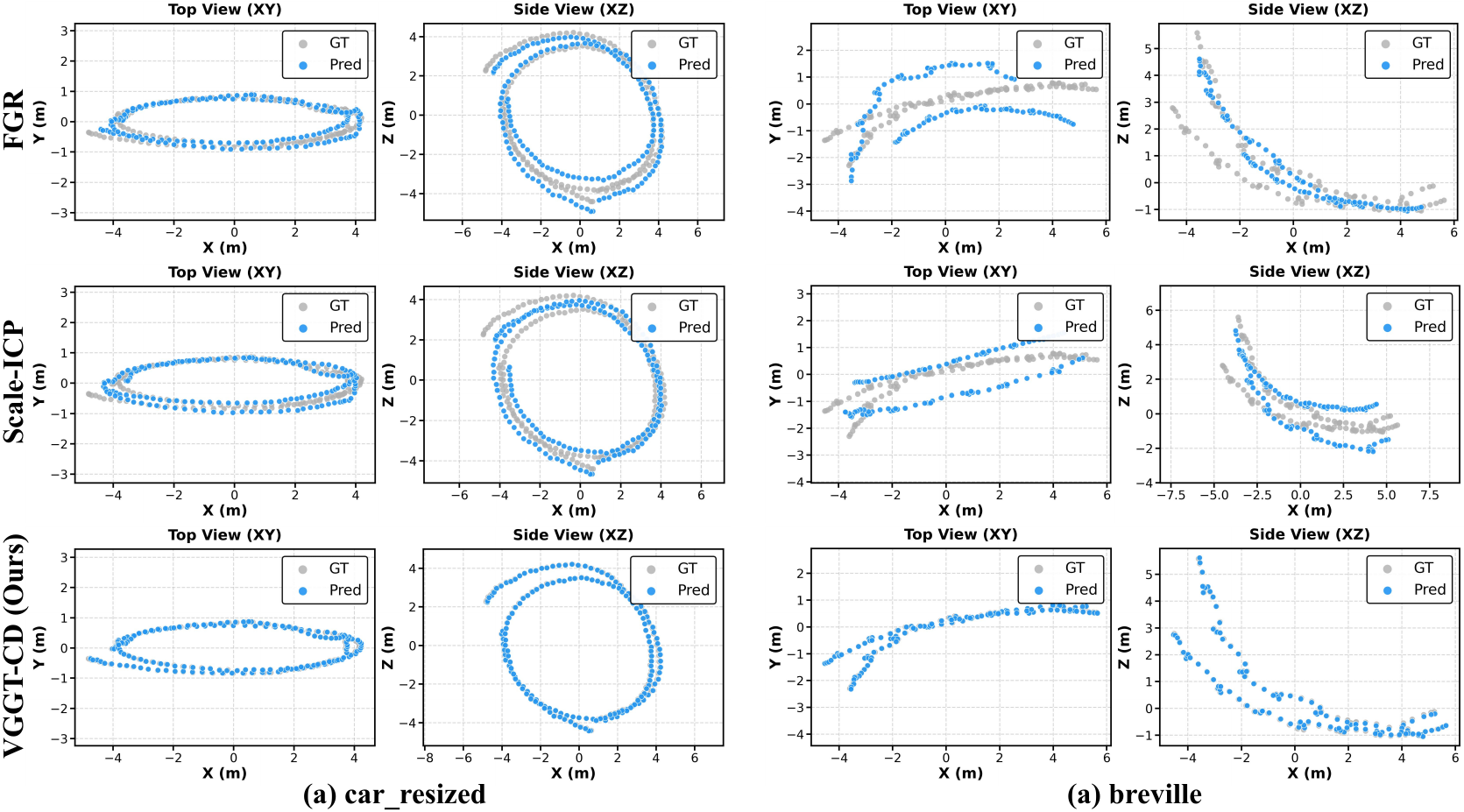}
  \caption{%
    Visualization of predicted camera trajectories versus ground truth
    on representative scenes.
    Each subplot shows the ground-truth trajectory (green)
    and the predicted trajectory (red) after cross-temporal alignment.
    \textbf{Top row:} VGGT + FGR.
    \textbf{Middle row:} VGGT + RANSAC+ICP.
    \textbf{Bottom row:} VGGT-CD (Ours).
    Baseline methods exhibit significant trajectory deviations
    due to scale ambiguity and outlier contamination
    from physical changes,
    whereas our method produces trajectories
    that closely overlap with the ground truth across all scenes.
  }
  \label{fig:trajectory}
\end{figure*}

\paragraph{\textbf{Analysis of the failure mode.}}
The experimental results reveal a systematic failure pattern
inherent to the ``independent reconstruction + registration'' paradigm.
When VGGT reconstructs each epoch independently,
the resulting point clouds reside in arbitrary coordinate frames
with unknown relative scale.
Post-hoc methods must then solve
a scale-aware $\mathrm{Sim}(3)$ registration problem
from point clouds that contain substantial dynamic changes.
These physical changes act as massive geometric outliers,
severely biasing the optimization.
Among the baselines, Scale+ICP mitigates the scale issue
through explicit scale estimation and performs well
on scenes with moderate scale variation
(\eg \textit{Community}, \textit{Ninja}),
yet still fails on scenes with larger ambiguity
(\eg \textit{Breville}, \textit{Dyson}).
In contrast, VGGT-CD resolves the scale ambiguity
\emph{before} dense registration via joint inference,
and then exploits the purification mechanism
to explicitly isolate static regions for refinement.
This is further corroborated by the trajectory visualization
in Fig.~\ref{fig:trajectory}:
baseline methods produce predicted trajectories
that deviate significantly from the ground truth
due to scale distortion and outlier contamination,
whereas the trajectories produced by VGGT-CD
closely overlap with the ground truth across all scenes.

\paragraph{\textbf{Registration efficiency.}}
VGGT-CD completes the registration stage
in 3--6\,s on average,
which is $6{\times}$--$255{\times}$ faster than all baselines
(Tables~\ref{tab:outdoor} and~\ref{tab:indoor}).
This efficiency stems from two factors:
(1)~the coarse stage requires only a single VGGT forward pass
on $2K{=}10$ keyframes,
and (2)~the fine stage involves
a single nearest-neighbor query and closed-form centroid computation
with no iterative optimization.
Even against the fastest baseline (Scale+ICP, 26\,s indoors on average),
VGGT-CD remains over $6{\times}$ faster,
while FGR and GeoTransformer
require hundreds to thousands of seconds
on scenes with large point clouds
(\eg 4608\,s for FGR on \textit{Ninja}).

\subsection{Qualitative Evaluation}\label{sec:qual}

\begin{figure*}[t]
  \centering
  \includegraphics[width=\linewidth]{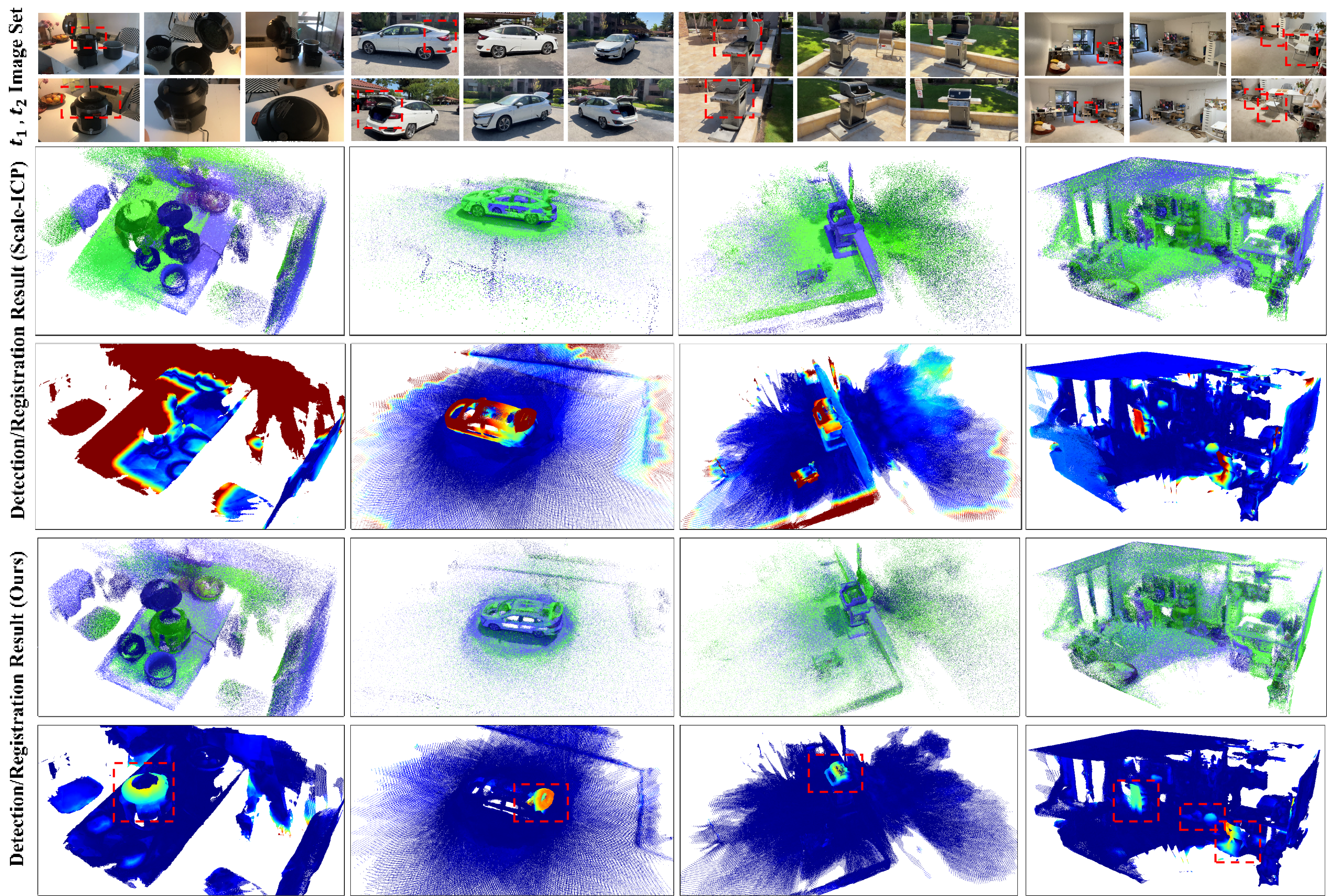}
  \caption{%
    Qualitative comparison on representative scenes.
    \textbf{Row 1:} input image sets from epoch $T_1$ and $T_2$.
    \textbf{Rows 2--3:} registration and change detection results of Scale-ICP, where misaligned static backgrounds produce widespread false positives that obscure genuine changes.
    \textbf{Rows 4--5:} registration and change detection results of VGGT-CD (Ours), where the static background is precisely aligned (blue), and genuine physical changes (\eg opened trunk, moved furniture, added objects) are cleanly highlighted with sharp spatial boundaries.
  }
  \label{fig:qualitative}
\end{figure*}

We further visualize the 3D change detection results
as spatial-distance heatmaps in Fig.~\ref{fig:qualitative}.
For each point in the aligned source cloud,
we color-code its nearest-neighbor distance to the target cloud:
blue indicates small distance (well-aligned static background),
while red indicates large distance (detected change).

For Scale-ICP,
the residual scale and pose misalignment
produces heatmaps dominated by false positives:
floors, walls, and ceilings that are physically unchanged
appear uniformly red,
rendering the true change regions
completely indistinguishable from registration artifacts.
In contrast,
VGGT-CD produces clean, high-contrast heatmaps.
The static background appears uniformly dark blue,
confirming precise metric alignment.
Genuine physical changes,
such as a newly opened car trunk,
a repositioned chair,
or a moved appliance,
are sharply highlighted in red
with clear spatial boundaries.
This qualitative evidence confirms that
high-precision static-background registration
is both necessary and sufficient
for accurate 3D change detection.

\subsection{Ablation Studies}\label{sec:ablation}

We conduct ablation experiments to validate
the contribution of each pipeline component.

\paragraph{\textbf{Effect of the Coarse Stage.}}
We first remove the sparse joint-inference module
(\ie ``W/o Stage~1''),
which forces the system to directly register
the independently reconstructed dense point clouds
without any scale prior.
As shown in Tables~\ref{tab:outdoor} and~\ref{tab:indoor}
(row ``VGGT Only''),
this configuration completely fails:
without the unified metric space established by joint inference,
the $\mathrm{Sim}(3)$ ambiguity
between the two epochs is unresolvable,
and ATE degrades to 0.3--3.4\,m.
We also verified a ``Fine Only'' configuration
that applies the purification mechanism
on top of the identity transform (without any coarse prior).
The resulting ATE remains comparable to the unregistered baseline
(\eg 3.37\,m on \textit{Car}, 0.73\,m on \textit{Grill}):
without a correct scale estimate,
the nearest-neighbor correspondences
between the two point clouds are meaningless,
and translation refinement alone cannot compensate.
These results demonstrate that the joint-inference mechanism
is \emph{indispensable} for establishing the metric-scale prior.

\begin{figure}[t]
  \centering
  \includegraphics[width=0.95\linewidth]{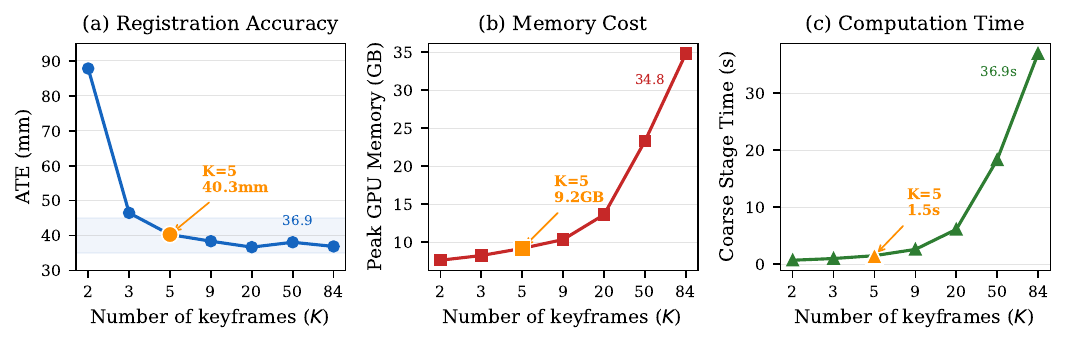}
  \caption{%
    \textbf{Effect of keyframe budget $K$ on the coarse stage.}
    (a)~ATE saturates around $K{=}5$;
    (b)~GPU memory grows super-linearly with $K$;
    (c)~computation time scales linearly.
    Orange markers highlight our default $K{=}5$.
  }
  \label{fig:ablation_kf}
\end{figure}

\paragraph{\textbf{Effect of keyframe budget ($K$).}}
We study the trade-off between the number of keyframes $K$
used in the coarse stage and the resulting
accuracy, memory cost, and computation time.
Fig.~\ref{fig:ablation_kf} reports results on \textit{Car}
with $K \in \{2, 3, 5, 9, 20, 50, 84\}$,
where $K{=}84$ corresponds to using all available frames
in joint inference.

Three clear trends emerge.
\textbf{(1)~Accuracy saturates early.}
ATE drops sharply from 87.8\,mm ($K{=}2$)
to 40.3\,mm ($K{=}5$),
then plateaus:
the marginal improvement from $K{=}5$
to full-sequence joint inference ($K{=}84$)
is only 3.4\,mm (8\%),
well within the noise floor of VGGT's single-pass reconstruction.
\textbf{(2)~Memory grows super-linearly.}
Peak GPU memory rises from 9.2\,GB ($K{=}5$)
to 34.8\,GB ($K{=}84$),
a $3.8{\times}$ increase
that would exceed the capacity of most consumer GPUs.
\textbf{(3)~Time grows linearly with $K$.}
The coarse stage takes 1.5\,s at $K{=}5$
versus 36.9\,s at $K{=}84$,
a $24.6{\times}$ slowdown
with negligible accuracy return.
These results validate our default choice of $K{=}5$
as a favorable operating point
that captures over 97\% of the full-sequence accuracy
at less than 4\% of the memory and time cost.

\paragraph{\textbf{Effect of the Fine Stage.}}
We retain the coarse alignment
but compare the result with and without
the dense purification and translation refinement of Stage~2.
Table~\ref{tab:ablation_fine} reports
the ATE (in mm) of both configurations
across four representative scenes and all keyframe budgets
from $K{=}2$ to full-sequence.

Two key properties are validated.
\textbf{(1)~Monotonicity guarantee.}
The improvement $\Delta$ is \emph{non-negative}
in every single cell of the table
(36 out of 36 configurations).
When the Fine Stage cannot improve upon the Coarse solution,
the residual self-check (Eq.~\ref{eq:selfcheck})
automatically reverts to the original translation,
ensuring zero degradation.
This occurs consistently for \textit{Car}
at $K {\geq} 20$, where the Coarse alignment
is already near-optimal.
\textbf{(2)~Adaptive recovery.}
The Fine Stage provides the largest correction
precisely when the Coarse quality is weakest.
At $K{=}2$, the improvement reaches
9.6\% on \textit{Car} (89.4$\to$80.8\,mm)
and 8.0\% on \textit{Grill} (40.2$\to$37.0\,mm).
On \textit{Mac}, where the scale factor is close to unity
($s {\approx} 0.97$) and the Coarse result
already achieves good alignment,
the Fine Stage still provides consistent
but smaller improvements of 0.3--1.4\%.
As $K$ increases and the Coarse alignment improves,
the Fine Stage gracefully reduces to a no-op on scenes
where it is not needed (\textit{Car}, \textit{Grill} at large $K$),
while continuing to contribute on \textit{Mac}
even at $K{=}\text{full}$ ($\Delta{=}1.1\%$).
This \emph{help-or-hold} behavior stems from
the decoupled design:
by locking $s$ and $\vR$ and only refining $\vt$
via centroid alignment on purified static correspondences,
the Fine Stage is a monotonically bounded operation
that adds less than 5\,s of overhead.

\begin{table*}[t]
  \centering
  \caption{%
    \textbf{Fine Stage ablation: ATE (mm) with and without Stage~2
    across keyframe budgets.}
    ``Coarse'' = Stage~1 only;
    ``Full'' = Stage~1 + Stage~2 (Ours);
    $\Delta$ = relative ATE improvement (\%).
    Deeper blue indicates larger improvement.
    All $\Delta \geq 0$: the residual self-check guarantees
    the Fine Stage never degrades the Coarse result.
  }
  \label{tab:ablation_fine}
  \setlength{\tabcolsep}{4pt}
  \renewcommand{\arraystretch}{1.15}
  \small
  \resizebox{0.8\textwidth}{!}{%
  \begin{tabular}{@{}l
    ccc
    ccc
    ccc
    ccc
    @{}}
    \toprule
    & \multicolumn{3}{c}{\textit{Car} (outdoor)}
    & \multicolumn{3}{c}{\textit{Mac} (indoor)}
    & \multicolumn{3}{c}{\textit{Grill} (outdoor)}
    & \multicolumn{3}{c}{\textit{Living Room} (indoor)} \\
    \cmidrule(lr){2-4} \cmidrule(lr){5-7} \cmidrule(lr){8-10} \cmidrule(lr){11-13}
    $K$
    & Coarse & Full & $\Delta$\,(\%)
    & Coarse & Full & $\Delta$\,(\%)
    & Coarse & Full & $\Delta$\,(\%)
    & Coarse & Full & $\Delta$\,(\%) \\
    \midrule
    2
    & 89.4 & 80.8 & \cellcolor{cvprblue!35} \textbf{9.6}
    & 63.0 & 62.1 & \cellcolor{cvprblue!12} 1.4
    & 40.2 & 37.0 & \cellcolor{cvprblue!35} 8.0
    & 533.6 & 520.3 & \cellcolor{cvprblue!18} 2.5 \\
    3
    & 45.9 & 43.1 & \cellcolor{cvprblue!30} 6.1
    & 59.2 & 59.0 & \cellcolor{cvprblue!3}  0.3
    & 32.3 & 31.2 & \cellcolor{cvprblue!22} 3.4
    & 594.2 & 579.9 & \cellcolor{cvprblue!18} 2.4 \\
    5
    & 39.3 & 39.0 & \cellcolor{cvprblue!8}  0.8
    & 57.9 & 57.7 & \cellcolor{cvprblue!3}  0.3
    & 39.1 & 38.6 & \cellcolor{cvprblue!12} 1.3
    & 454.7 & 452.4 & \cellcolor{cvprblue!5}  0.5 \\
    7
    & 39.3 & 38.8 & \cellcolor{cvprblue!12} 1.3
    & 63.6 & 62.7 & \cellcolor{cvprblue!12} 1.4
    & 38.2 & 37.8 & \cellcolor{cvprblue!10} 1.0
    & 486.4 & 481.4 & \cellcolor{cvprblue!10} 1.0 \\
    9
    & 36.2 & 36.1 & \cellcolor{cvprblue!3}  0.3
    & 59.7 & 59.4 & \cellcolor{cvprblue!5}  0.5
    & 38.6 & 38.4 & \cellcolor{cvprblue!5}  0.5
    & 468.4 & 463.4 & \cellcolor{cvprblue!10} 1.1 \\
    15
    & 38.3 & 38.2 & \cellcolor{cvprblue!3}  0.3
    & 63.7 & 63.0 & \cellcolor{cvprblue!10} 1.1
    & 36.0 & 35.9 & \cellcolor{cvprblue!3}  0.3
    & 448.4 & 448.4 & \cellcolor{white}       0.0 \\
    20
    & 35.4 & 35.4 & \cellcolor{white}       0.0
    & 62.6 & 61.9 & \cellcolor{cvprblue!10} 1.1
    & 32.9 & 32.9 & \cellcolor{white}       0.0
    & 457.0 & 457.0 & \cellcolor{white}       0.0 \\
    50
    & 36.4 & 36.4 & \cellcolor{white}       0.0
    & 62.6 & 61.8 & \cellcolor{cvprblue!12} 1.3
    & 35.8 & 35.8 & \cellcolor{white}       0.0
    & 475.8 & 471.0 & \cellcolor{cvprblue!10} 1.0 \\
    \rowcolor{gray!8}
    full
    & 35.7 & 35.7 & \cellcolor{white}       0.0
    & 62.1 & 61.4 & \cellcolor{cvprblue!10} 1.1
    & 34.1 & 34.1 & \cellcolor{white}       0.0
    & 475.8 & 471.0 & \cellcolor{cvprblue!10} 1.0 \\
    \bottomrule
  \end{tabular}%
  }
\end{table*}

\section{Conclusion}\label{sec:conclusion}
We have presented \ours{},
a training-free, coarse-to-fine pipeline
for 3D change detection from multi-view images.
By exploiting the joint-inference capability
of the VGGT foundation model,
our Coarse Stage resolves cross-temporal scale ambiguity
with only a handful of keyframes,
while the Fine Stage refines alignment
through confidence-guided purification
and decoupled translation estimation,
backed by a residual self-check
that guarantees monotonic improvement.
Experiments on an 11-scene benchmark
demonstrate consistent gains over all baselines,
with up to 59\% ATE reduction
and $6{\times}$--$255{\times}$ faster registration,
yielding clean, high-contrast 3D change maps
without any task-specific training.

\paragraph{Limitations and future work.}
Our method inherits the reconstruction quality
of the underlying VGGT model;
in scenes with extreme depth range
or very sparse viewpoint overlap,
the absolute registration error remains higher
than in well-covered scenes.
When the inter-epoch scale variation is small
($s \approx 1.0$),
direct scale-aware ICP can occasionally
achieve comparable accuracy (\eg \textit{Ninja}),
suggesting that an adaptive registration strategy
could further improve robustness.
Integrating semantic priors for finer-grained change categorization
and extending to more than two temporal epochs
are promising directions for future work.


\end{document}